\pdfoutput=1
\documentclass[final]{cvpr}

\usepackage{times}
\usepackage{epsfig}
\usepackage{graphicx}
\usepackage{amsmath}
\usepackage{amssymb}
\usepackage{blindtext}
\usepackage{multirow}

\usepackage[pagebackref=true,breaklinks=true,colorlinks,bookmarks=false]{hyperref}

\def\cvprPaperID{****} 
\def\confYear{CVPR 2021}

\begin{document}
\title{PP-YOLOv2: A Practical Object Detector}

\author{
		 Xin Huang$^{1,2}$, Xinxin Wang$^1$, Wenyu Lv$^1$, Xiaying Bai$^1$, Xiang Long$^1$ \\
		 Kaipeng Deng$^1$, Qingqing Dang$^1$, Shumin Han$^1$, Qiwen Liu$^1$, Xiaoguang Hu$^1$  \\
		 Dianhai Yu$^1$, Yanjun Ma$^1$, Osamu Yoshie$^2$ \\
		{\tt\small koushin@toki.waseda.jp, 
			\{wangxinxin08, lvwenyu01, baixiaying, longxiang
		}\\
		{\tt\small
			 dengkaipeng, dangqingqing, hanshumin, liuqiwen, huxiaoguang
		} \\
		{\tt\small
			 yudianhai, mayanjun02\}@baidu.com, yoshie@waseda.jp
		} \\
		$^1$Baidu Inc.  $^2$Waseda University\\
	}

\maketitle
\begin{abstract}
   Being effective and efficient is essential to an object detector for practical use. To meet these two concerns, we comprehensively evaluate a collection of existing refinements to improve the performance of PP-YOLO while almost keep the infer time unchanged. This paper will analyze a collection of refinements and empirically evaluate their impact on the final model performance through incremental ablation study. Things we tried that didn't work will also be discussed. By combining multiple effective refinements, we boost PP-YOLO's performance from 45.9\% mAP to 49.5\% mAP on COCO2017 test-dev. Since a significant margin of performance has been made, we present PP-YOLOv2. In terms of speed,  PP-YOLOv2 runs in 68.9FPS at 640x640 input size. Paddle inference engine with TensorRT, FP16-precision and batch size = 1 further improves PP-YOLOv2's infer speed, which achieves 106.5 FPS. Such a performance surpasses existing object detectors with roughly the same amount of parameters (\emph{i.e.}, YOLOv4-CSP, YOLOv5l). Besides, PP-YOLOv2 with ResNet101 achieves 50.3\% mAP on COCO2017 test-dev. Source code is at \url{https://github.com/PaddlePaddle/PaddleDetection}.
\end{abstract}

\section{Introduction}
\begin{figure}[t]
		\begin{center}
			\includegraphics[width=0.95\linewidth]{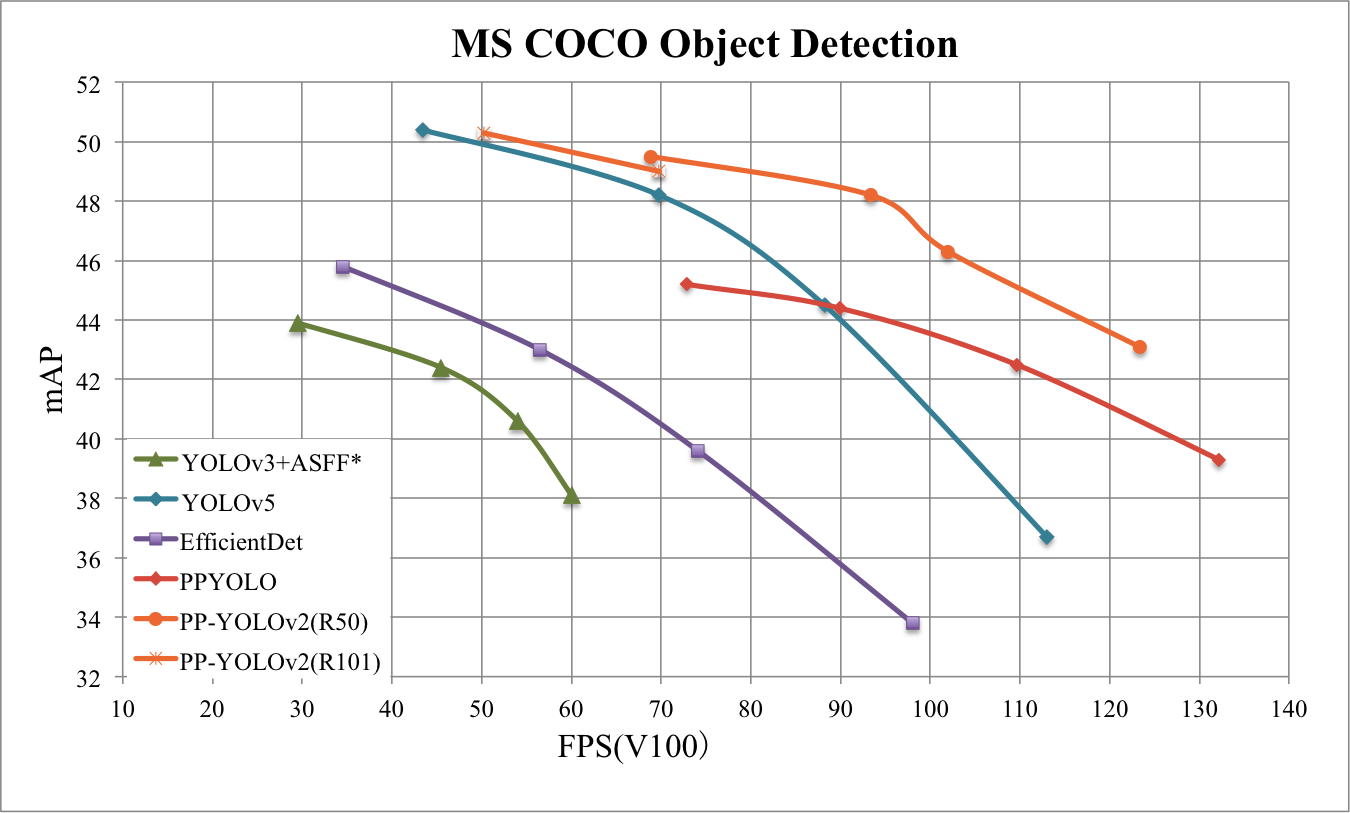}
		\end{center}
		\caption{Comparison of the proposed PP-YOLOv2 and other object detectors.  With a similar FPS, PP-YOLOv2 outperforms YOLOv5l by 1.3\% mAP. Besides, when we replace PP-YOLOv2's backbone from ResNet50 to ResNet101, PP-YOLOv2 achieves comparable performance with YOLOv5x while it is 15.9\% faster than YOLOv5x. The data is recorded in Table \ref{tab2}.}
		\label{fig:fps}
\end{figure}
Object detection is a critical component of various real-world applications such as self-driving cars, face recognition, and person re-identification. In recent years, the performance of object detectors has been rapidly improved with the rise of deep convolutional neural networks (CNNs)~\cite{simonyan2014very, he2016deep, hu2018squeeze}. Although, recent works focus on novel detection pipeline (\emph{i.e.}, Cascade RCNN~\cite{cai2018cascade} and HTC~\cite{chen2019hybrid}), sophisticated network architecture design (DetectoRS~\cite{qiao2020detectors} and CBNET~\cite{liu2020cbnet}) push forward the state-of-the-art object detection approaches, YOLOv3~\cite{redmon2018YOLOv3} is still one of the most widely used detector in industry. Because, in various practical applications, not only the computation resources are limited, but also the software support is insufficient. Without necessary technique support, two stage object detector(\emph{e.g.} Faster RCNN~\cite{ren2015faster}, Cascade RCNN~\cite{cai2018cascade}) may excruciatingly slow. Meanwhile, a significant gap exists between the accuracy of YOLOv3 and two stage object detectors. Therefore, how to improve the effectiveness of YOLOv3 while maintaining the inference speed is an essential problem for practical use. To simultaneously satisfy two concerns, we add a bunch of refinements that almost not increase the infer time to improve the overall performance of the PP-YOLO~\cite{long2020pp}. To note that, although a huge number of approaches claim to improve object detector's accuracy independently, in practice, some methods are not effective when combined. Therefore, practical testing of combinations of such tricks is required. We follow the incremental manner to evaluate their effectiveness one by one. All our experiments are implemented based on PaddlePaddle\footnote{https://github.com/PaddlePaddle/Paddle}~\cite{ma2019paddlepaddle}.

\begin{figure*}[t!]
	\centering
	\includegraphics[width=1.0\textwidth]{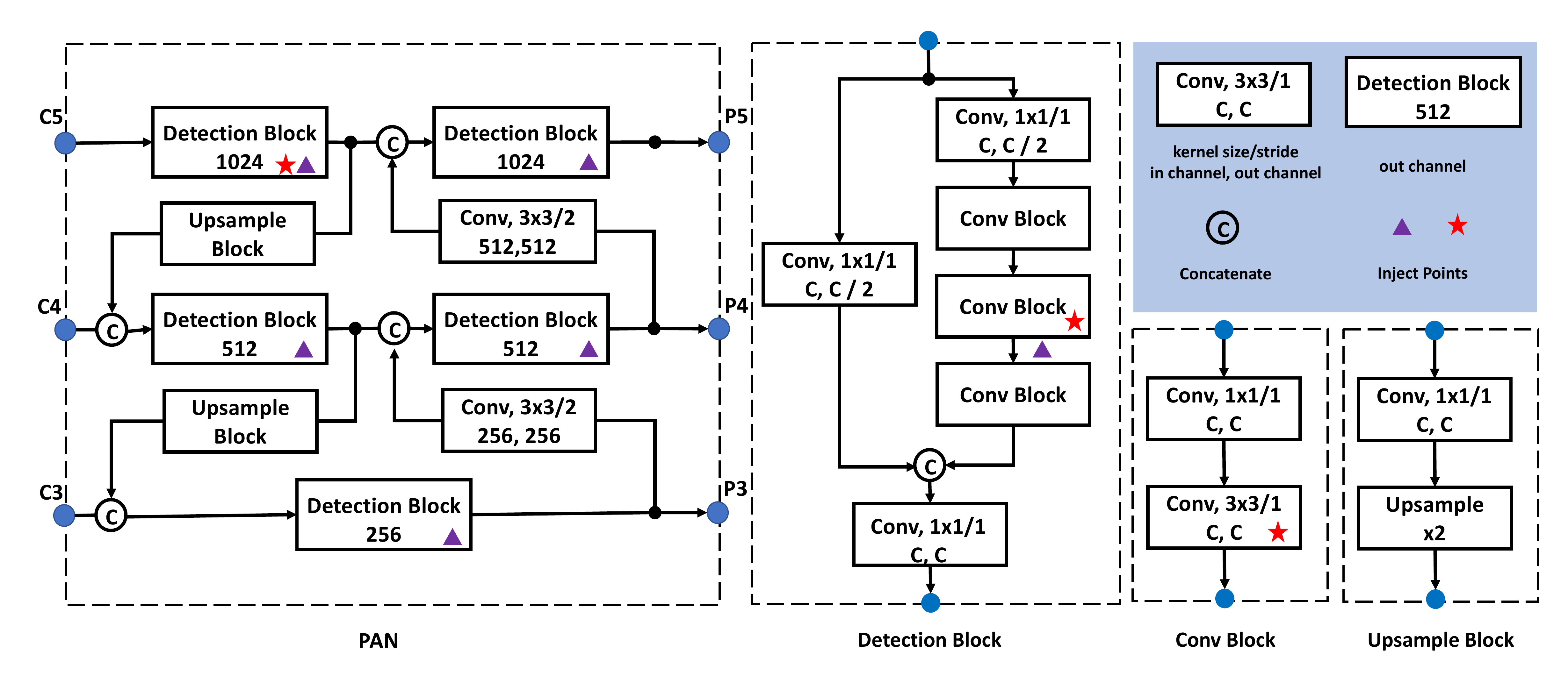}\\
	\caption{The architecture of PP-YOLOv2's detection neck.}
	\label{fig:pan}
\end{figure*}

In fact, this paper is more like a TECH REPORT, which tells you how to build PP-YOLOv2 step by step. Theoretical justification of the failure cases is also involved. To this end, we achieve a better balance between effectiveness (49.5\% mAP) and efficiency (69 FPS), surpassing existing robust detectors with roughly the same amount of parameters such as YOLOv4-CSP ~\cite{wang2020scaledYOLOv4} and YOLOv5l\footnote{https://github.com/ultralytics/yolov5}. Hopefully, our experience in building PP-YOLOv2 can help developers and researchers to think deeper in implementing object detectors for practical applications.

\section{Revisit PP-YOLO}
In this section, we will perform the implementation of our baseline model specifically. 

\noindent \textbf{Pre-Processing.} Apply Mixup Training~\cite{zhang2017mixup} with a weight sampled from $Beta({\alpha}, {\beta})$ distribution where $\alpha = 1.5,  \beta=1.5.$ Then, RandomColorDistortion, RandomExpand, RandCrop and RandomFlip are applied one by one with probability 0.5. Next, Normalize RGB channels by subtracting 0.485, 0.456, 0.406 and dividing by 0.229, 0.224, 0.225, respectively. Finally,  The input size is evenly drawn from [320, 352, 384, 416, 448, 480, 512, 544, 576, 608].

\noindent \textbf{Baseline Model.} Our baseline model is PP-YOLO which is an enhanced version of YOLOv3. Specifically, it first replaces the backbone to ResNet50-vd\cite{he2019bag}. After that a total of 10 tricks which can improve the performance of YOLOv3 almost without losing efficiency are added to YOLOv3 such as Deformable Conv~\cite{dai2017deformable}, SSLD~\cite{cui2021selfsupervision}, CoordConv~\cite{liu2018intriguing}, DropBlock~\cite{ghiasi2018dropblock}, SPP~\cite{he2015spatial} and so on. The architecture of PP-YOLO is presented in the paper~\cite{long2020pp}. 

\noindent \textbf{Training Schedule.}  On COCO \textit{train2017}, the network is trained with stochastic gradient descent (SGD) for 500K iterations with a minibatch of 96 images distributed on 8 GPUs. The learning rate is linearly increased from 0 to 0.005 in 4K iterations, and it is divided by 10 at iteration 400K and 450K, respectively. Weight decay is set as 0.0005, and momentum is set as 0.9. Gradient clipping is adopted to stable the training procedure.

\begin{table*}[t!]
		\centering
		\begin{tabular}{l|l|l|l|l|l|l}
			\hline
			& \textbf{Methods}  & \textbf{mAP} & \textbf{Parameters} & \textbf{GFLOPs} & 
			\textbf{infer time} & \textbf{FPS} \\ 
			\hline
			\hline
			A & PP-YOLO  & 45.1 &45 M & 45.1 & 13.7 ms\dag& 72.9  \\ 
			\hline
			\hline
			B & A + PAN + MISH  & 47.1 & 54 M & 52.0 & 14.0 ms& 71.4 \\ 
			C & B + input size 640 & 47.7 &54 M  & 57.6 & 14.5 ms& 68.9\\ 
			D & C + Larger input size & 48.3 &54 M & 57.6 & 14.5 ms& 68.9\\ 
			E & D + IoU Aware Branch & 49.1 &54 M & 57.6 & 14.5 ms& 68.9 \\ 
			\hline
		\end{tabular}
		\caption{The ablation study of refinements on the MS-COCO minival split. ”\dag” indicates the result includes bounding box decode time(1\textasciitilde2ms).  }
		\label{tab1}
\end{table*}

\section{Selection of Refinements}

\vspace{0.3cm}
\noindent \textbf{Path Aggregation Network.} Detecting objects at different scales is a fundamental challenge in object detection. In practice, a detection neck is developed for building high-level semantic feature maps at all scales. In PP-YOLO, FPN is adopted to compose bottom-up paths. Recently, several FPN variants have been proposed to enhance the ability of pyramid representation. For example, BiFPN~\cite{tan2020efficientdet}, PAN~\cite{liu2018path}, RFP~\cite{qiao2020detectors} and so on. We follow the design of PAN to aggregate the top-down information. The detailed structure of PAN is shown in Fig.~\ref{fig:pan}. 

\vspace{0.3cm}
\noindent \textbf{Mish Activation Function.} Mish activation function~\cite{misra2019mish} has been proved effective in many practical detectors, such as YOLOv4 and YOLOv5. They adopt the mish activation function in the backbone. However, we prefer to use pre-trained parameters because we have a powerful model which achieves 82.4\% top-1 accuracy on ImageNet. To keep the backbone unchanged, we apply the mish activation function in the detection neck instead of the backbone. 

\vspace{0.3cm}
\noindent \textbf{Larger Input Size.} Increasing the input size enlarges the area of objects. Thus, information of the objects on a small scale will be preserved easier than before. As a result, performance will be increased. However, a larger input size occupies more memory. To apply this trick, we need to decrease batch size. To be more specific, we reduce the batch size from 24 images per GPU to 12 images per GPU and expand the largest input size from 608 to 768. The input size is evenly drawn from [320, 352, 384, 416, 448, 480, 512, 544, 576, 608, 640, 672, 704, 736, 768].

\vspace{0.3cm}
\noindent \textbf{IoU Aware Branch.} In PP-YOLO, IoU aware loss is calculated in a soft weight format which is inconsistent with the original intention. Therefore, we apply a soft label format. Here is the IoU aware loss:
\begin{align}
	loss = -t * \log(\sigma(p)) - (1 - t) * \log(1 - \sigma(p))
\end{align}
where $t$ indicates the IoU between the anchor and its matched ground-truth bounding box, $p$ is the raw output of IoU aware branch, $\sigma(\cdot)$ refers to the sigmoid activation function. To note that only positive samples' IoU aware loss is computed. By replacing the loss function, IoU aware branch works better than before.

\section{Experiments}

\subsection{Dataset}
COCO~\cite{lin2014microsoft} is a widely used benchmark in the field of object detection. In this work, we train all our models on the COCO \textit{train2017} which consists of 118k images across 80 classes. For evaluation, we evaluate our results on the COCO \textit{minival} which consists of 5k testing images. Our evaluation metric also follows the standard COCO style mean Average Precision (mAP). 

\subsection{Ablation Studies}
In this subsection, we present the effectiveness of each module in an incremental manner. Results are shown in Table \ref{tab1}, where infer time and FPS only consider the influence of the model in FP32-precision which does not include result decoder and NMS following YOLOv4\cite{bochkovskiy2020YOLOv4}.

\vspace{0.3cm}
\noindent \textbf{A.} 
First of all, we follow the original design of PP-YOLO to build our baseline. Since the heavy pre-processing on the CPU slows down the training, we decrease the images per GPU from 24 to 12. Reducing batch size drops mAP by 0.2\%. Training settings are described in section 2 entirely. 

\vspace{0.3cm}
\noindent \textbf{A $\rightarrow$ B.} The first refinement with a positive effect on PP-YOLO that we found was PAN. To stable the training process, we add several skip connections to our PAN module. The detailed structure of PAN is shown in Fig.~\ref{fig:pan}. We can see that PAN and FPN are a group of symmetrical structures. When we perform it with Mish, it boosts the performance from 45.1\% mAP to 47.1\% mAP. Although model B is slightly slower than model A, such a significant gain promotes us to adopt PAN in our final model.  For
more details, please refer to our code.

\vspace{0.3cm}
\noindent \textbf{B $\rightarrow$ C.} Since the input size of YOLOv4 and YOLOv5 during evaluation is 640, we increase training and evaluation input size to 640 to build a fair comparison. The performance increases 0.6\% mAP. 

\vspace{0.3cm}
\noindent \textbf{C $\rightarrow$ D.} Keep increasing the input size should benefit more. However, it is impossible to use Larger Input Size and Larger Batch Size together. We train the model D with 12 images per GPU and Larger Input Size. It increases the mAP by 0.6\% which brings more gains than Larger Batch Size. Therefore, we choose Larger Input Size in the final practice. The input size is evenly drawn from [320, 352, 384, 416, 448, 480, 512, 544, 576, 608, 640, 672, 704, 736, 768].

\vspace{0.3cm}
\noindent \textbf{D $\rightarrow$ E.} In the training phase, the modified IoU aware loss performs better than before. In the former version, the value of IoU aware loss will drop to 1e-5 in hundreds of iterations during training. After we modified the IoU aware loss, its value and the value of IoU loss are in the same order of magnitude, which is reasonable. After using this strategy, the mAP of model E increases to 49.1\% without any loss of efficiency.

\begin{table*}[h]
		\centering
		\resizebox{1.0\textwidth}{!}{
			\begin{tabular}{l|l|c|cc|cccccc}
				\hline
				\multirow{2}{*}{\textbf{Method}} & \multirow{2}{*}{\textbf{Backbone}} & \multirow{2}{*}{\textbf{Size}} &\multicolumn{2}{c|}{\textbf{FPS (V100)}} &
				\multirow{2}{*}{\textbf{AP}} & \multirow{2}{*}{\textbf{AP$_{50}$}} & \multirow{2}{*}{\textbf{AP$_{75}$}} & \multirow{2}{*}{\textbf{AP$_S$}} & \multirow{2}{*}{\textbf{AP$_M$}} & \multirow{2}{*}{\textbf{AP$_L$}}\\	
				\cline{4-5} 
				& & & \textbf{w/o TRT} & \textbf{with TRT} & & & & & &\\			
				\hline
				\hline
				YOLOv3 + ASFF* \cite{Liu2019Learning} & Darknet-53 & 320 &  60   &- & 38.1\% & 57.4\% & 42.1\% & 16.1\% & 41.6\% & 53.6\% \\
				YOLOv3 + ASFF* \cite{Liu2019Learning}& Darknet-53 & 416 &  54   & - &40.6\% & 60.6\% & 45.1\% & 20.3\% & 44.2\% & 54.1\% \\
				YOLOv3 + ASFF* \cite{Liu2019Learning}& Darknet-53 & 608 &  45.5   & - &42.4\% & 63.0\% & 47.4\% & 25.5\% & 45.7\% & 52.3\% \\
				YOLOv3 + ASFF* \cite{Liu2019Learning} & Darknet-53 & 800 & 29.4   &- & 43.9\% & 64.1\% & 49.2\% & 27.0\% & 46.6\% & 53.4\% \\
				\hline
				EfficientDet-D0~\cite{tan2020efficientdet} & Efficient-B0  & 512 &  98.0$^+$ & - & 33.8\% & 52.2\% & 35.8\% & 12.0\% & 38.3\% & 51.2\% \\
				EfficientDet-D1~\cite{tan2020efficientdet} & Efficient-B1 & 640 &  74.1$^+$  &- & 39.6\% & 58.6\% & 42.3\% & 17.9\% & 44.3\% & 56.0\% \\
				EfficientDet-D2~\cite{tan2020efficientdet} & Efficient-B2 & 768 & 56.5$^+$  &- & 43.0\% & 62.3\% & 46.2\% & 22.5\% & 47.0\% & 58.4\% \\
				EfficientDet-D2~\cite{tan2020efficientdet} & Efficient-B3 & 896 & 34.5$^+$  &- & 45.8\% & 65.0\% & 49.3\% & 26.6\% & 49.4\% & 59.8\% \\
				\hline
				YOLOv4~\cite{bochkovskiy2020YOLOv4} & CSPDarknet-53 & 416 & 96  &164.0$^*$ & 41.2\% & 62.8\% & 44.3\% & 20.4\% & 44.4\% & 56.0\% \\
				YOLOv4~\cite{bochkovskiy2020YOLOv4} & CSPDarknet-53 & 512 & 83  &138.4$^*$  & 43.0\% & 64.9\% & 46.5\% & 24.3\% & 46.1\% & 55.2\% \\
				YOLOv4~\cite{bochkovskiy2020YOLOv4} & CSPDarknet-53 & 608  & 62  &105.5$^*$  & 43.5\% & 65.7\% & 47.3\% & 26.7\% & 46.7\% & 53.3\% \\
				\hline
				YOLOv4-CSP~\cite{wang2020scaledYOLOv4} & Modified CSPDarknet53 & 512 & 97   & - & 46.2\%  & 64.8\% & 50.2\% & 24.6\% & 50.4\% & 61.9\% \\
				YOLOv4-CSP~\cite{wang2020scaledYOLOv4} &  Modified CSPDarknet53 & 640 & 73  &  - & 47.5\% & 66.2\% & 51.7\% & 28.2\% & 51.2\% & 59.8\% \\
				\hline
				YOLOv5s & - & 640 & 113$^*$ &  - & 36.7\% & 55.4\% & - & - & - & - \\
				YOLOv5m & - & 640 & 88.2$^*$ & - & 44.5\% & 63.1\% & - & - & - & - \\
				YOLOv5l & - & 640 & 69.8$^*$ & - & 48.2\%  & 66.9\% & - & - & - & - \\
				YOLOv5x& - & 640 & 43.4$^*$ & - & 50.4\% & 68.8\% & - & - & - & - \\
				\hline
				PP-YOLO~\cite{long2020pp} & ResNet50-vd-dcn & 320 & 132.2\dag &  242.2\dag & 39.3\% & 59.3\% & 42.7\% & 16.7\% & 41.4\% & 57.8\% \\
				PP-YOLO~\cite{long2020pp} & ResNet50-vd-dcn & 416 & 109.6\dag & 215.4\dag & 42.5\% & 62.8\% & 46.5\% & 21.2\% & 45.2\% & 58.2\% \\
				PP-YOLO~\cite{long2020pp} & ResNet50-vd-dcn & 512 & 89.9\dag & 188.4\dag & 44.4\%  & 64.6\% & 48.8\% & 24.4\% & 47.1\% & 58.2\% \\
				PP-YOLO~\cite{long2020pp}& ResNet50-vd-dcn & 608 & 72.9\dag &  155.6\dag & 45.9\% & 65.2\% & 49.9\% & 26.3\% & 47.8\% & 57.2\% \\
				\hline
				\hline
				PP-YOLOv2 & ResNet50-vd-dcn & 320 & 123.3 & 152.9 & 43.1\% & 61.7\% & 46.5\% & 19.7\% & 46.3\% & 61.8\% \\
				PP-YOLOv2 & ResNet50-vd-dcn & 416 & 102 & 145.1 & 46.3\% & 65.1\% & 50.3\% & 23.9\% & 50.2\% & 62.2\% \\
				PP-YOLOv2 & ResNet50-vd-dcn & 512 & 93.4 &  141.2 & 48.2\% & 67.1\% & 52.7\% & 27.7\% & 52.1\% & 62.1\% \\
				PP-YOLOv2 & ResNet50-vd-dcn & 608 & 72.1 & 109.9 & 49.2\% & 68.0\% & 54.1\% & 29.9\% & 52.8\% & 61.5\% \\
				PP-YOLOv2 & ResNet50-vd-dcn & 640 & 68.9 & 106.5 & 49.5\% & 68.2\% & 54.4\% & 30.7\% & 52.9\% & 61.2\% \\
				\hline
				PP-YOLOv2 & ResNet101-vd-dcn & 512 & 69.8 & 116.8 & 49.0\%  & 67.8\% & 53.8\% & 28.7\% & 53.0\% & 63.5\% \\
				PP-YOLOv2& ResNet101-vd-dcn & 640 & 50.3 &  87.0 & 50.3\% & 69.0\% & 55.3\% & 31.6\% & 53.9\% & 62.4\% \\
				\hline
			\end{tabular}
		}
		\caption{Comparison of the speed and accuracy of different object detectors on the MS-COCO (test-dev 2017). We compare the results with batch size = 1, without tensorRT (w/o TRT) or with tensorRT(with TRT).
			Results marked by "+" are updated results from the corresponding official code base.
			Results marked by "*" are test in our environment using official code and model. ”\dag” indicates the result includes bounding box decode time(1\textasciitilde2ms). The backbone of YOLOv5 has not been named yet, so we leave it blank.
		}
		\label{tab2}
\end{table*}

\subsection{Comparison With Other State-of-the-Art Detectors}
Comparison of the results on MS-COCO test split with other state-of-the-art object detectors is shown in Figure \ref{fig:fps} and Table \ref{tab2}. We compare our method with YOLOv4-CSP and YOLOv5l because they have roughly the same amount of parameters as our model. It clearly shows that PP-YOLOv2 outperforms these two methods. With a similar FPS, PP-YOLOv2 outperforms YOLOv4-CSP by 2\% mAP and surpasses YOLOv5l by 1.3\% mAP. Besides, when we replace PP-YOLOv2's backbone from ResNet50 to ResNet101, PP-YOLOv2 achieves comparable performance with YOLOv5x while it is 15.9\% faster than YOLOv5x. Therefore, we can draw a conclusion that compared with other state-of-the-art methods, our PP-YOLOv2 has certain advantages in the balance of speed and accuracy. 

Moreover, PP-YOLOv2 is implemented based on PaddlePaddle. As a deep learning framework, PaddlePaddle not only supports model implementation but also pays attention to model deployment. With official support, adapting TensorRT for PP-YOLOv2 is much easier than other detectors. Specifically, the Paddle inference engine with TensorRT, FP16-precision, and batch size = 1 further improves PP-YOLOv2's infer speed. The speed-up ratios for PP-YOLOv2(R50) and PP-YOLOv2(R101) are 54.6\% and 73\%, respectively.
\section{Things We Tried That Didn't Work}
Since it takes about 80 hours for training PP-YOLO with 8 V100 GPUs on COCO \textit{train2017}, we involve COCO \textit{minitrain}~\cite{sheshkus2019houghnet} to speed up our analysis on ablation studies. COCO \textit{minitrain} is a subset of the COCO \textit{train2017}, containing 25K images. On COCO \textit{minitrain}, the total iterations is 90K. We divide the learning rate by 10 at iteration 60k.  Other settings are the same as training on COCO \textit{train2017}. 

 We tried lots of stuff while we were working on PP-YOLOv2. Some of them have a positive effect on COCO \textit{minitrain} while hinders the performance when training on COCO \textit{train2017}. Due to the inconsistency, someone may doubt the experimental conclusion on COCO \textit{minitrain}.The reason why we use COCO \textit{minitrain} is that we want to seek refinements with universal features, which means that they should be useful on different scale datasets. It is also important to figure out the reason why they failed. Therefore, we discuss some of them in this section. 

\noindent \textbf{Cosine Learning Rate Decay.} Different from linear step learning rate decay, cosine learning rate decay is exponentially decaying the learning rate. Therefore, the change of the learning rate is smooth, which benefits the training process. We follow the formula in Bag of Tricks \cite{he2019bag} to set learning rate at each epoch. Although cosine learning rate decay achieves better performance on COCO \textit{minitrain}, it is sensitive to hyper-parameters such as initial learning rate, the number of warm up steps, and the ending learning rate. We tried several sets of hyper-parameters. However, we didn't observe a positive effect on COCO \textit{train2017} eventually.

\noindent \textbf{Backbone Parameter Freezing.} When fine-tuning the ImageNet pre-trained parameters on downstream tasks, freezing parameters in the first two stages is a common practice. Since our pre-trained ResNet50-vd is much powerful than others(82.4\% Top1 accuracy versus 79.3\% Top1 accuracy), we are more motivated to adopt this strategy. On COCO \textit{minitrain}, parameter freezing brings 1mAP gain, however, on COCO \textit{train2017} it decreases mAP by 0.8\% . A possible reason for the inconsistency phenomena was speculated to be the different sizes of the two training sets. COCO \textit{minitrain} is a fifth of COCO \textit{train2017}. The ability to generalization of parameters that are trained on a small dataset may be worse than pre-trained parameters.

\noindent \textbf{SiLU Activation Function.} We tried using SiLU~\cite{wang2018lidar} instead of Mish in detection neck. This increases 0.3\%  mAP on COCO \textit{minitrain} but drops 0.5\%  mAP on COCO \textit{train2017}. We are not sure about the reason.
\section{Conclusions}
This paper presents some updates to PP-YOLO, which forms a high-performance object detector called PP-YOLOv2. PP-YOLOv2 achieves a better balance between speed and accuracy than other famous detectors, such as YOLOv4 and YOLOv5. In this paper, we explore a bunch of tricks and show how to combine these tricks on the PP-YOLO detector and demonstrate their effectiveness. Moreover, with PaddlePaddle's official support, the gap between model development and production deployment is narrowed.  We hope this paper can help developers and researchers get better performance in practical scenes.

\section{Acknowledgements}
This work was supported by the National Key Research and Development Project of China (2020AAA0103500).
{\small

\bibliographystyle{ieee_fullname}
\bibliography{ppyolov2}
}
\end{document}